# Generative Model-Driven Synthetic Training Image Generation: An Approach to Cognition in Rail Defect Detection

Rahatara Ferdousi[1], Chunsheng Yang[2], M. Anwar Hossain[3], Fedwa Laamarti[1,4], M. Shamim Hossain[5], Abdulmotaleb El Saddik[1,4]

[1]The University of Ottawa, Ottawa, Ontario, Canada

[2]National Research Council Canada, Ottawa, Ontario, Canada

[3]School of Computing, Queen's University, Kingston, Ontario, Canada

[4]Computer Vision Department, Mohamed bin Zayed University of Artificial Intelligence, UAE

[5] Department of Software Engineering, College of Computer and Information Sciences, King Saud University, Saudi Arabia.

Corresponding author: Rahatara Ferdousi (e-mail: rferd068@uottawa.ca); M. Shamim Hossain (e-mail: mshossain@ksu.edu.sa.)

## Abstract

Recent advancements in cognitive computing, with the integration of deep learning techniques, have facilitated the development of intelligent cognitive systems (ICS). This is particularly beneficial in the context of rail defect detection, where the ICS would emulate human-like analysis of image data for defect patterns. Despite the success of Convolutional Neural Networks (CNN) in visual defect classification, the scarcity of large datasets for rail defect detection remains a challenge due to infrequent accident events that would result in defective parts and images. Contemporary researchers have addressed this data scarcity challenge by exploring rule-based and generative data augmentation models. Among these, Variational Autoencoder (VAE) models can generate realistic data without extensive baseline datasets for noise modeling. This study proposes a VAE-based synthetic image generation technique for rail defects, incorporating weight decay regularization and image reconstruction loss to prevent overfitting. The proposed method is applied to create a synthetic dataset for the Canadian Pacific Railway (CPR) with just 50 real sam-ples across five classes. Remarkably, 500 synthetic samples are generated with a minimal reconstruction loss of 0.021. A Visual Transformer (ViT) model underwent fine-tuning using this synthetic CPR dataset, achieving high accuracy rates (98%-99%) in classifying the five defect classes. This research offers a promising solution to the data scarcity challenge in rail defect detection, showcasing the potential for robust ICS development in this domain.

A comprehensive video presentation is available for viewing via this link: Video Presentation on SharePoint.

Please find a practical demonstration of the data generator, as well as the source code, through the following GitHub link: GenAI For Goods on GitHub.

# 1 Introduction

In recent years, cognitive computing has advanced through the integration of deep learning, aiming to mimic human thought processes in a computational context [1] [2]. This has prompted the development of Intelligent Cognitive Systems (ICS). One possible application of ICS is the detection and monitoring of rail defects, where the system is capable of understanding the visual patterns of defective parts much like a real-life quality inspector would do [3]. The overview of how ICS employs cognitive computing to transform human expertise through Artificial Intelligence (AI) is illustrated in Fig. 1.

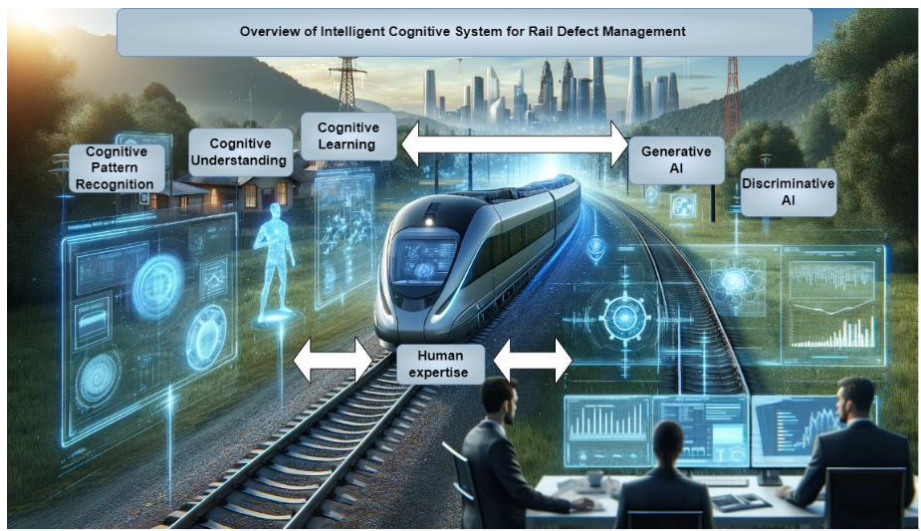

Figure 1: The ICS for rail defect detection mimics human expertise by using Generative AI to understand defect patterns and generate synthetic images for accurate classification, while Discriminative AI employs deep learning to recognize patterns and classy defect.

Despite the success of Convolutional Neural Networks (CNNs) in visual clas-sification [4], CNN models and their variations often suffer from generalizability issues in rail defect classification [5]. This is particularly true when the clas-sification models are often trained and evaluated on an insufficient number of sample images [6].

One of the reasons behind incorporating limited samples in the existing approach is the scarcity of original defect samples or the challenge of collecting such samples [7]. This is attributed to the rare occurrence of accidents, due to high quality-controlling and inspection measures in railway system [8] [9]. In addition, the quality of images that would characterize rail defects can deteri-orate due to dynamically changing backgrounds and lighting conditions while collecting the data [10]. This data scarcity issue ultimately limits the ability of ICS to 'learn' from real-world examples to achieve human-like accuracy and decision-making capabilities [11].

There are two common approaches to handling the data scarcity issue for training a classifier: i) Fixed-rule based [12] and ii) Data-driven models [13] [14]. The traditional fixed rule-based data augmentation (e.g., zooming, rotation, etc.) may not adequately capture the intricate patterns and variations present in actual defect images [15].

The Data-driven models on the other hand have become popular in defect detection due to their ability to fit distributions accurately [15]. Among these models, Generative Adversarial Network (GAN) models have been used in var-ious use cases where the reconstructed images are supposed to be realistic but significantly distinguishable from the real images [16]. However, GANs are sen-sitive to data distribution and require a significant amount of data for effective noise modeling [17]. By contrast, Variational Auto Encoder (VAE)-based gen-erative models are designed to acquire a probabilistic mapping of the data [18]. When trained on limited data, VAEs can still effectively model the statistical properties and patterns of the available samples within the latent space [19]. The learned probabilistic mapping allows VAEs to generate synthetic samples that closely resemble the real data, even with a small number of training exam-ples [20].

Therefore, in this research, we propose a VAE-based synthetic data gener-ation approach. The model takes defective images as input, encodes high-level features (e.g., edge, shape, intensity), learns the latent distribution of the de-fects, and finally reconstructs synthetic samples from the learned patterns. As the VAE model learns the latent space from a limited sample size [21], it may encounter constraints leading to the reconstruction of unrealistic or erroneous samples [22]. Therefore, we update the regularization technique of the tradi-tional VAE model by adding weight decay to the reconstruction training loss to subdue the overfitting issue. Furthermore, we outlined the algorithm for syn-thetic dataset generation using VAE model. The algorithm outputs the desired number of reconstructed samples for each class.

We conducted an experiment using the main CPR Dataset [7], which in-cludes 50 samples belonging to three normal classes and two defect classes. First, we evaluated the VAE model in terms of its ability to reconstruct images and tested it on the overall loss. Second, we measured the performance of the fine-tuned vision transformer-based defect classifier, which was trained using the CPR Synthetic dataset. The CPR Synthetic dataset contains 500 images, including 450 reconstructed images generated from the 50 original images of the main CPR dataset. We found that our modified regularization technique reduces loss compared to the traditional regularization technique of the VAE model. Furthermore, the proposed VAE approach showed better qualitative re-sults. It improved the distribution of the latent space and dealt with issues of wrong image reconstruction caused by overfitting. During the second phase of the evaluation, we achieved high scores ranging from 98% to 99% in terms of test accuracy, training accuracy, F-measure, precision, and recall. Our findings suggest that the VAE model can be beneficial in addressing the data collection challenge for rarely occurring samples like defects.

The rest of the paper is organized as follows. In Section 2, we outline the existing approach for synthetic data generation to identify the gaps and require-ments for the existing approach. Section 3 details the proposed VAE-based approach and Synthetic data generation algorithm incorporating the VAE. In Section 4, we present the data, experiments, and results. We discuss the limi-tations and viable future works in Section 5. Finally, we conclude this study in Section 6.

## 2 Literature Review

The challenge of data scarcity in training classifiers is a significant hurdle in automating defect inspection for large-scale transportation systems. Synthetic image generation has emerged as a popular solution in the industry for ad-dressing this issue [23]. We have reviewed the existing literature and found that researchers have extensively explored various approaches to handle this challenge, which we outline in the following sections.

### 2.1 Fixed-rule based approach

The fixed-ruled-based data augmentation typically involves geometric transfor-mations (e.g., zooming, rotation, cropping, scaling) or basic image processing (e.g., color contrasting) [24]. However, the generated defects from a fixed-rule-based approach often lack randomness to mimic real defects [24, 25]. Slight variations in existing defects may not be fully captured or adequately described by the preset rules, leading to degraded quality of the generated defects and limitations in providing appropriate synthetic data for the defect classifier [25].

### 2.2 Data-driven approach

Data-driven synthetic defect generation methods that learn directly from exist-ing data are another approach [15]. From the literature on defect classification, we identified two candidate data-driven approaches: i) Generative Adversarial Networks (GAN) [26] and ii) Variational Autoencoders (VAE) [27]. We con-ducted a comparison between GAN and VAE to gain an understanding of the effectiveness and suitability of the GAN and VAE approaches specifically for synthetic data generation for rail defects.

Table 3 contrasts the two data-driven generative models, making it clear that VAEs offer a higher degree of control and flexibility, especially for cases where data variation might be limited, and where preservation of semantic information in generated data is crucial.

Table 1: Comparative Table: GAN vs. VAE for Defect Data Generation [26,27]

| Subject | GAN | VAE |
|---|---|---|
| Latent Space Control | No Controlling of the latent space | Has control over the latent space. |
| Noise Modeling | Complex and Data Sensitive | Not Required. |
| Variation | Not suitable for limited variation | Can learn latent distribution from limited variation |
| Semantic Representations of generated data | Not preserved | Preserved and modifiable |

## 2.3 Related work

The use of GenAI for synthetic data generation is a novel approach to address the issue of limited datasets. In this section, we provide a summary of the few existing studies that are related to this topic. The application and methods in these works are tabulated in Table 2

Table 2: Comparison of Papers on Synthetic Data Generation

| Paper | Method Used | Application |
|---|---|---|
| [12] | GANs | Surface defect detection |
| [28] | DT-GAN | Defect synthesis |
| [13] | Defect-GAN | Defect inspection |
| [14] | VAEs | Compound property prediction |
| [29] | Generative AI models | Healthcare |
| [23] | reviewed Various methods (including decision trees, deep learning techniques, and iterative proportional fitting) | Various applications |

The authors in [12] propose a framework for data augmentation by creating synthetic images using GANs. The synthetic images are used for training classification algorithms, improving the performance of CNN for the classification of surface defects. This approach is similar to the one used in [13] and [28], but it specifically focuses on surface defects. In the context of railway defect detection, this approach could be particularly useful as surface defects are the most common type of railway defect.

The study in [28] presents Defect Transfer GAN, a framework that can learn to depict different types of defects using different backdrop products. It can also apply styles particular to defects to create realistic representations of defects. When compared to cutting-edge picture synthesis approaches, it generates more diverse defects and has higher sample fidelity. This study, along with the study in [13], provides a more specialized approach to defect synthesis compared to the general synthetic data augmentation approach presented in [12].

The authors in Defect-GAN [13] present an automated defect synthesis net-work that generates realistic and diverse defect samples for training accurate and robust defect inspection networks. It shows that Defect-GAN can synthe-size various defects with diversity and fidelity.

The authors [14] suggests a technique to enhance the accuracy of machine learning models in predicting chemical properties. This is achieved by inte-grating supplementary data on associated molecular descriptors into the learnt representations of variational autoencoders. This work stands out as it focuses on improving the performance of VAEs, which is a different approach compared to the GAN-based methods used in the other papers. Although the applica-tion domain is not relevant to defect data generation, the findings of the paper helped to understand the role of latent-space-based VAE data generation for limited samples.

Similarly, the study in [29] examines generative AI models for creating real-istic, anonymized patient data for research and training. It explores synthetic data generation strategies for a use case where data collection is challenging due to privacy concerns.

Another comprehensive systematic review in [23] examines existing studies that employ machine learning models for generating synthetic data. It explores different machine learning methods, with particular emphasis on neural net-work approachs and deep generative models, and addresses privacy and fairness concerns related to synthetic data generation.

Although the existing literature provides valuable insights into synthetic data generation using GANs and VAEs, there remains a distinct gap in apply-ing these methods specifically to the challenges of railway defect classification, especially in the context of limited defect data, which we have outlined in the following section.

## 2.4 Gaps and Requirements

Most of the existing work has well-documented the generation of synthetic im-ages. However, there's an apparent gap in terms of evaluating and implementing these techniques in a complete pipeline, specifically for railway defect classifi-cation. A methodology that encompasses data generation, model training, and defect classification, tested in a real-world railway setting, is still yet to be explored.

After reviewing the recent research that proposed GAN and VAE for synthetic data generation, we found that GAN is limited to scenarios when the noise in the data is not properly modeled with small or less diverse training data [23]. In such cases, GANs face challenges in accurately capturing the un-derlying data distribution due to incomplete representation of noise patterns, leading to model collapse [26].

Therefore, we opted for VAE due to its ability to learn latent space from limited variation and fewer samples [27]. However, whenever a model learns from fewer samples there is a risk of overfitting issues [30]. In general, reconstruction loss is measured throughout the training process of VAE to keep the difference between the original and reconstructed image low [17].

However, if any model parameters are abnormally prioritized by the model, it will cause an overfitting issue resulting in a wrong or low-quality reconstruc-tion image, which is a concern when creating training data for the defect classi-fier [31]. The overfitting of the VAE model needs to be regularized while training the model.

Therefore, in this paper, we propose a VAE-based synthetic data generation method with a modified regularization mechanism to address these issues. In the following sections, we detail the methodology for synthetic data generation, model training, and defect classification. In addition, we rigorously test our approach for real railway defect datasets to address the identified research gap.

## 3 Method

In this section, we present the proposed VAE-based method for synthetic rail defect data generation.

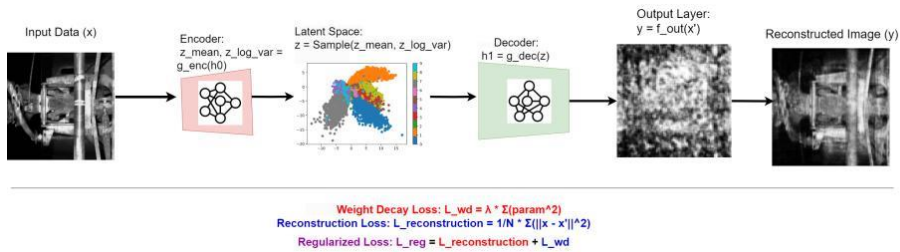

Figure 2: Proposed VAE approach for Rail Defect Image Reconstruction

Overall, the input layer takes the original images as input, then the encoder CNN captures its hidden representation, such as shape, texture, and other high-level features. After that, this encoding is used to learn a latent space. From this latent space, a pattern or distribution is learned by the decoding neural network. Finally, the output is pre-processed to reconstruct an image of the original image. Throughout this process, a regularization technique is applied to minimize the reconstruction loss and penalize large weights, mitigating the risk of overfitting. In Fig. 2, we illustrate the proposed VAE approach for rail defect image reconstruction. The detail of the proposed approach is presented in the following sections.

### 3.1 Encoder

In the context of a Variational Autoencoder (VAE), the hidden representation (h0) refers to the output of the encoder network after processing the input defect images. It captures the compressed representation of the input data (e.g., texture, shape, pattern, and edge). The hidden layers of the encoder CNN transform the hidden representation (h0) to obtain the parameters that define the probability distribution of the latent space. Let us consider that there

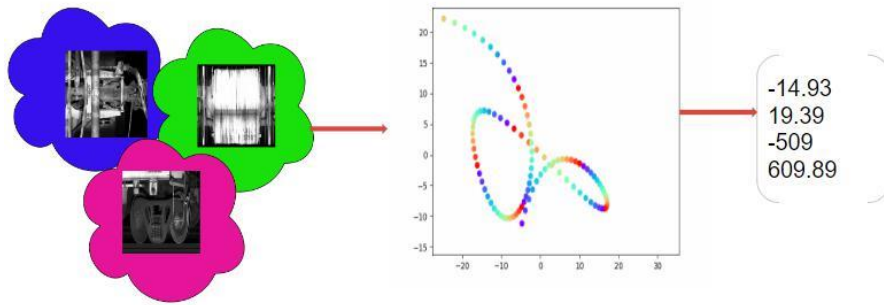

Figure 3: Illustraion of Encoding Process for Rail Defect Generation

are three different types of defects - bottom defect, side defect, and top defect. The VAE encoder-

- first, divides these input images into three regions, each represented by a different color in Figure 3.

- then processes each region separately and extracts features using convo-lutional layers.

- after that, it maps the extracted features of each region into a lower-dimensional vector (latent space).

- then the encoder generates multiple latent space vectors for each input image, allowing for probabilistic encoding. The variation in latent space allows the VAE to learn a meaningful and continuous latent space for better defect representation and image generation.

## 3.2 Latent Space

The latent space captures the essence of the input image. In the latent space, images that are similar or share common characteristics are expected to be closer to each other. Conversely, images that are dissimilar or have distinct features are expected to be farther apart. This property of latent space ensures that images with the same defect type are encoded close to each other in the latent space.

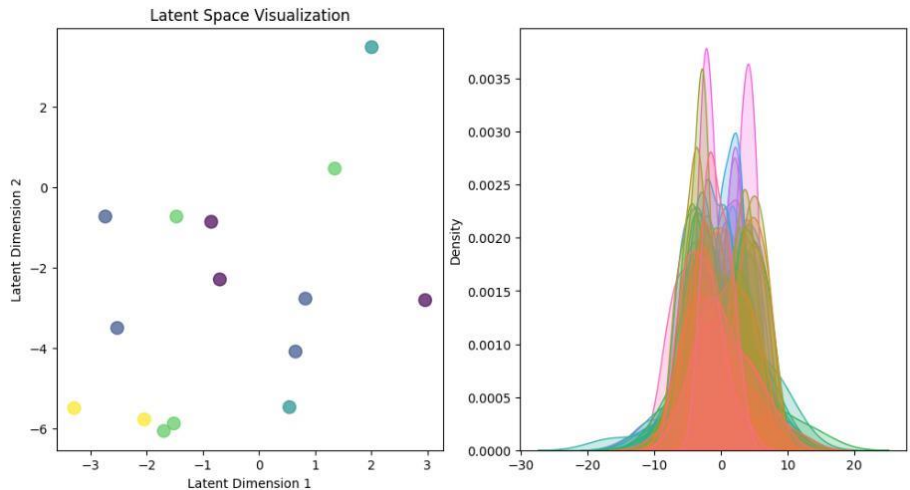

Figure 4: Example of a learned distribution of the latent space: On the left side, the same color represents samples of the same class, indicating that similar data points are clustered together in the latent space. The probability distribution on the right side shows the density and spread of the learned representations for different defect classes.

As demonstrated in Fig. 4, neighboring points in the latent space correspond to similar representations or characteristics in the generated data. By exploring different points in the latent space, such as sampling along a straight line or within a region, synthetic samples can be generated through gradual transitions or variations in their features. This property enables the VAE to perform smooth interpolation, allowing for the generation of new data points with controlled changes or combinations of attributes.

## 3.3 Decoder

The Decoder of VAE maps the latent space back to the data space. Unlike the encoder, the decoder takes the latent vector (z), maps it to the hidden representation (h1) and then generates the reconstructed defect image. To break it down, the decoder network (g dec)

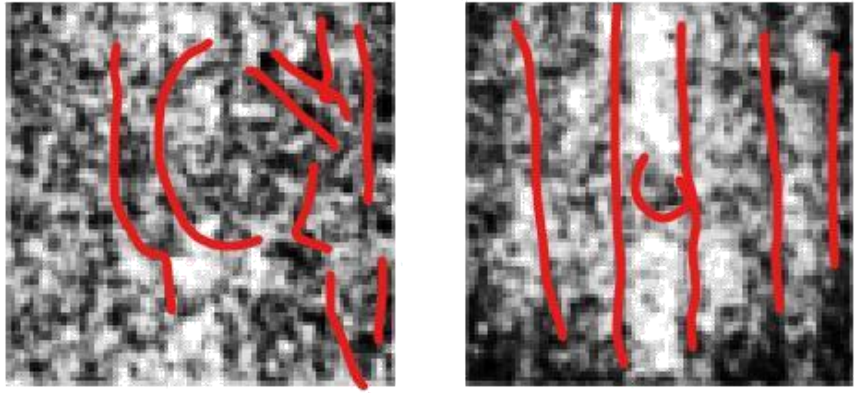

Figure 5: Example out of a decoder learned patterns (e.g., shapes) of a defective part. The red lines are to highlight the pattern in these images.

- takes the sampled latent vector (z) as input.
- then it maps the sampled latent vector (z) to a hidden representation (h1). Here h1, serves as a compressed and abstract representation of the input data, capturing important features such as texture, shape, pattern, and edges.
- after that, the decoder network uses several hidden layers to transform the hidden representation (h1) to generate the learned pattern as demon-strated in Fig. 5

Finally The reconstructed output $(x')$ is further processed by the output layer function (f out) to obtain the final output (y), which is the reconstructed Images in desired representation.

### 3.4 Regularization Technique

In general, a VAE-based method tries to keep the reconstruction loss low while learning. In our case, we propose adding weight decay and reconstruction loss to regularize the training. The loss to activate the regularization are computed as given in the following equations.

- Reconstruction Loss:

$$L_{reconstruction} = \frac{1}{N} \| x_{train} - x'_{train} \|^2$$

- Measures the discrepancy between the input data (x) and the reconstructed output $(x')$.

- A VAE-based method aims to minimize this loss during training to improve the quality of the reconstructions.

- Weight Decay Loss:

$$L_{wd} = \lambda \, param^2$$

- Penalizes large parameter values to control model complexity and avoid overfitting.
- Example: If $\lambda = 0.001$ and a parameter value is 2, the weight decay loss contribution for that parameter is $0.001 \times 2^2 = 0.004$.

- Combined Regularization:

$$\text{Total Loss} = L_{reconstruction} + L_{wd}$$

- We propose to add weight decay ($L_{wd}$) along with the reconstruction loss to regulate the training process.
- By adjusting the regularization weight $\lambda$, we can control the relative importance of the reconstruction term and overfitting prevention.

The combination of the reconstruction loss and weight decay in the regularization process can mitigate the trade-off between model complexity and generalization capability. In scenarios with limited training data, the proposed technique allows the VAE to effectively learn the underlying probability distribution from the available samples, avoiding the risk of overfitting, and enhancing its capacity to generalize to new data.

### 3.5 VAE-based Synthetic Data Generation Algorithm

The proposed Algorithm 1, employs a trained VAE model to reconstruct images of defects. The algorithm takes original images, the desired number of recon-structed images for each original image, and the total number of defect classes as inputs. Then it reconstructs the specified quantity of synthetic images for each class.

The variation in defect classes, represented as a set C is leveraged by the algorithm so that each defect class $\in$ C contributes to a diverse set of images. By iterating over each defect classes the proposed algorithm ensures variability in the defect characteristics of generated images.

In addition, it includes a randomization step to avoid repeating the same patterns while reconstruction process. The Algorithm 1 activates the VAE in evaluation mode to allow its encoder to approximate the posterior distribution $q_\phi(z|x')$. Latent codes are sampled from this distribution, creating a compact representation: $z \sim q_\phi(z|x')$.

The complexity of the algorithm for synthetic image generation is influenced by several factors including the number of defect classes (|C|), the total number of original images across all classes (N), the VAE's latent dimension (d), and the number of synthetic images generated per original image

| Algorithm 1: VAE-based Synthetic Image Generator |
|---|

Require: dataset path: Path to the original dataset
Require: def ect classes: List of defect classes
Require: reconstructed path: Path to save the reconstructed and original images
Require: vae model: Variational Autoencoder model
Require: num images per sample: Number of synthetic images per original sample

1: Create a directory at reconstructed path to save the reconstructed and original images
2: Set vae model to evaluation mode
3: Define image transformation steps: grayscale conversion, resizing, and conversion to tensor
4: for each defect class in defect classes do
5:    Set class path as dataset path/defect class
6:    Create a directory at reconstructed path/defect class + "-reconstructed" to store the reconstructed images
7:    Get the list of image files in class path
8:    Shuffle the image files randomly
9:    for each image in the shuffled image files do
10:       Load and pre-process the image
11:       Reshape the image to match the expected input shape of the VAE model
12:       Reconstruct the image using the VAE model
13:       Save the reconstructed image and the original image in the reconstructed directory.
14:    end for
15: end for
16: return Reconstructed and original images saved in the reconstructed path directory

(num images per sample). Moreover, a reconstruction process involves both encoding and decoding with time complexity of O(d). For an entire dataset, the total time complexity is calculated as:

$$O(|C| \cdot N \cdot num\_images\_per\_sample \cdot d) \tag{1}$$

For a large number of defect classes, the latent spaces can be high-dimensional. Consequently, the memory, model approach, and hardware resources may im-pact the scalability of the algorithm. Therefore, optimizations such as parallel processing and GPU acceleration could be considered to enhance the algorithm's scalability and performance for using the proposed algorithm for large-scale datasets.

## 4  Evaluation

This section details the dataset, system configuration, evaluation metric, and qualitative and quantitative results of the synthetic data generation approach. In addition, we present a further evaluation of a ViT defect classifier [32] trained on the generated synthetic dataset.

### 4.1  Data

The original dataset used in this study is the CPR main dataset. In Fig. 6 we demonstrate sample images from each of 5 defect classes representing the visual characteristics of the defects. It is spectacular from this illustration these images are hard to characterize manually by textures or color. However, the shape of the patterns is distinguishable.

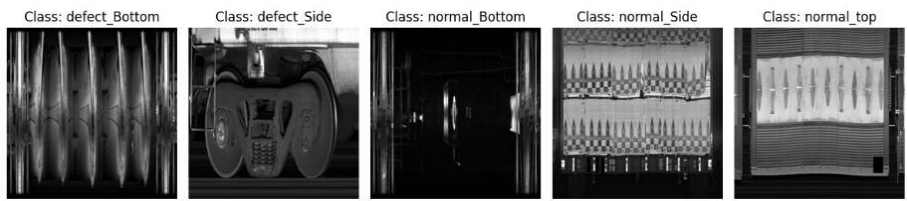

Figure 6: Example CPR original image samples for different classes.

Moreover, the bar diagram demonstrates that the sample distribution per class was not reasonable for training a CNN model. The distribution of samples per class in Fig. 7 demonstrates that it is infeasible to train a multi-class defect classifier, even with a transfer learning model. Because after splitting the dataset for training and testing there were significantly low samples (6-8 images in each class).

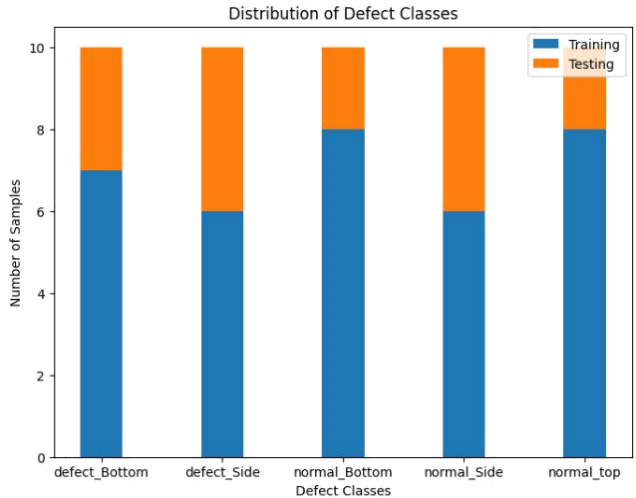

Figure 7: 80:20 split of training and testing data for the CPR main dataset

## 4.2 Evaluation Metrics

We employed a combination of various loss and accuracy measures, as presented in Table 3, to evaluate the quality of the generated outputs by the proposed VAE-based approach, as well as the classification accuracy of the ViT model.

The selected loss functions and performance measures facilitate specific purposes in evaluating the VAE generator and ViT model. The Reconstruction Loss is used to measure the similarity between the original and reconstructed images by the generative models, while the KLD demonstrates the ability of the VAE model to capture underlying data distributions. We considered widely used performance measures, including Precision, Recall, and F1-Score, for as-sessing the performance of the ViT defect classifier in terms of false positives and false negative scores for each class.

## 4.3 Evaluation of VAE model

In this section, we present two types of comparisons for the proposed approach. First, we compare the performance between the traditional and proposed approach. Second, we compare the qualitative output of the proposed VAE approach with two other contemporary synthetic data generation approaches.

### 4.3.1 Comparison between traditional and proposed VAE model

- Latent space analysis The plots in Fig. 8 represent the learning process of the latent space for different classes. Compared to the traditional VAE's latent space, the proposed VAE's latent space demonstrates a clear separation between different classes because the points in Fig. 8a are closer for the same classes. This outcome indicates that the model has learned to encode the input data in a way that captures the underlying class-specific features.

Table 3: Loss Functions and Classifier Performance Metrics.
TP = True positive, FP= False Positive, TN= True Negative, FN= False Negative

| Metrics | Description / Formula |
|---|---|
| **1. Training Loss** | |
| Reconstruction Loss + Weight Decay | $L_{reconstruction} + L_{wd}$ |
| Weight Decay Loss ($L_{wd}$) | $L_{wd} = \lambda \sum (param^2)$ |
| Reconstruction Loss ($L_{reconstruction}$) | $L_{reconstruction} = \frac{1}{N} \sum (\|x_{train} - x'_{train}\|^2)$ |
| **2. Test Loss** | |
| Reconstruction Loss + Kullback-Leibler Divergence | $L_{reconstruction} + KLD$ |
| Reconstruction Loss ($L_{reconstruction}$) | $L_{reconstruction} = \frac{1}{N} \sum (\|x_{test} - x'_{test}\|^2)$ |
| Kullback-Leibler Divergence (KLD) | $KLD = \sum P(x) \log \frac{P(x)}{Q(x)}$ |
| P (True distribution) | |
| Q (Learned distribution) | |
| **3. Classifier Performance Metrics** | |
| Precision | $\frac{TP}{TP+FP}$ |
| Recall | $\frac{TP}{TP+FN}$ |
| F1-Score | $\frac{2 \cdot Precision \cdot Recall}{Precision + Recall}$ |
| Test Accuracy | Accuracy on test data |

The t-SNE (t-Distributed Stochastic Neighbor Embedding) is performed to obtain these plottings [31]. t-SNE performs dimensionality reduction of the latent vectors from their original high-dimensional space to a lower-dimensional space to compress the representation of the original image. Here, the resulting vectors have two dimensions representing the pattern of the encoded images in the latent space. The better one group of (same color) points is linearly separable from the other groups of data by a straight line, the better the model learns about the variation in patterns of different classes [31].

However, some overlaps can be observed in both of the latent spaces, which suggests that the model needs improvement to create distinct representa-tions for each class. Overlapping points between classes indicate that the model struggles to differentiate between certain instances the classes have similarities, which is one of the limitations of the CPR dataset. Because some of the samples in this dataset from different classes are visually iden-tical.

- Training Loss The provided code generates a comparative summary of training loss over 100 epochs for both the traditional and proposed VAE

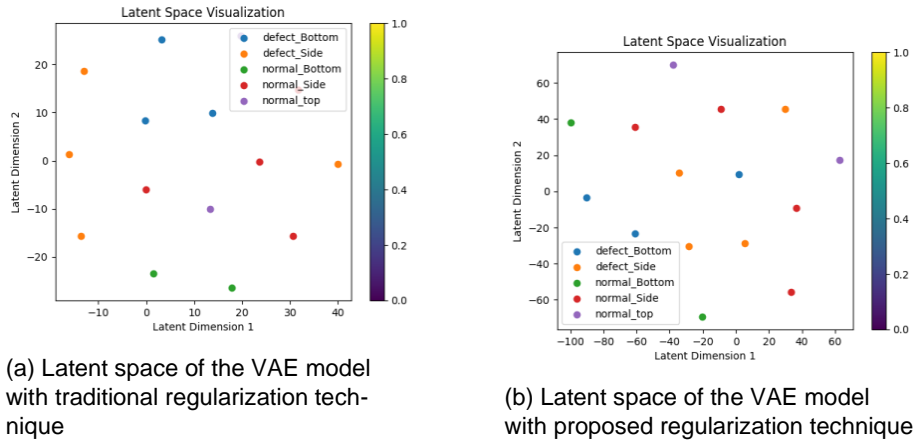

(a) Latent space of the VAE model with traditional regularization technique

(b) Latent space of the VAE model with proposed regularization technique

Figure 8: The latent vectors represent the encoded latent space representations of the images, and the classes are represented by different colors. The x-axis represents the first dimension of the reduced latent vectors, and the y-axis represents the second dimension. Each point in the scatter plot

approaches. The loss values for each approach are plotted on the y-axis against the epoch numbers on the x-axis.

Although both of the curves exhibit a consistent downward trend, there is a subtle difference in terms of consistency in performance.

The Fig. 9, demonstrates that the traditional VAE approach demonstrates a relatively high initial loss value of around 12.13 and experiences a rapid decrease in loss over the first few epochs. In contrast, the line for the proposed approach starts with a lower initial loss value of approximately 0.1087 and steadily reduces the loss throughout the training process. At the end of the 100th epoch, both of the approaches achieve a significantly lower final loss value of about 0.0081 after 100 epochs. However, the high loss and fluctuation of the performance by the traditional approach are clearly an indication of an overfitting issue, which has also been reflected in the qualitative output, which we discuss next.

- Quality of generated output We illustrate the output of the VAE model with the traditional regularization technique in Fig. 10a and the out of the updated regularization technique in Fig. 10a in order to observe the quality of the synthetic images.

It is evident from the output images that the VAE model with the tra-ditional regularization technique provides better resolution and correct output in comparison with the proposed VAE model with the updated regularization technique. To elaborate, having a closer look at the images in Fig. 10a, we can observe that the first constructed image in the list is blurry and the features of the images are hardly visible. In addition,

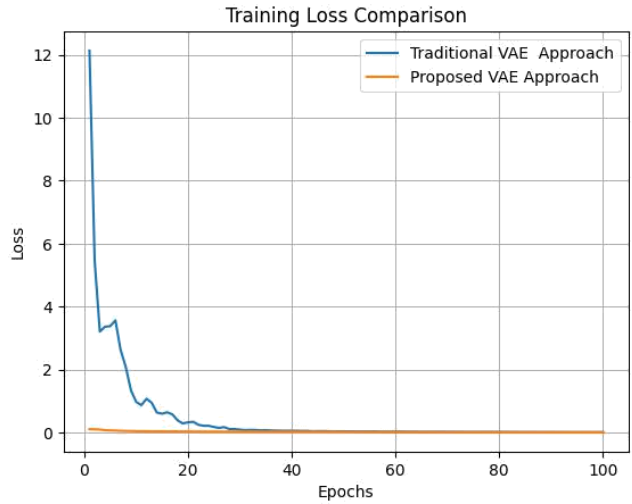

Figure 9: Training loss comparison graph of VAE model with traditional regularization technique VS VAE model with proposed regularization technique

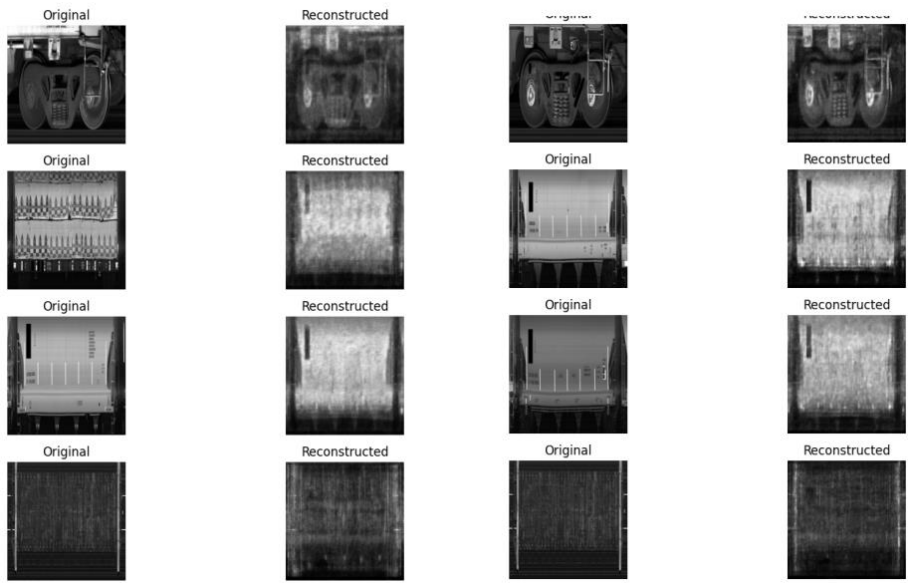

(a) Synthetic images generated by the VAE model with traditional regularization technique

(b) Synthetic images generated by the VAE model with proposed regularization technique

the second generated image in Fig. 10a is totally a different image. As a consequence, training a classifier with such poor-quality images leads to lower accuracy in the model ranging from 58%-61%.

- Test Loss We obtained the test loss by combining the reconstruction loss and KLD to observe the difference between the original and reconstructed images. While generating samples using the traditional VAE we obtained low loss values, nearly converging to 0. By contrast, with the incorporation of weight decay into our training the test loss was reduced from an initial value of 0.030 to a significantly improved 0.021. Interestingly, the training loss was

### 4.3.2 Comparison between the proposed and existing data genera-tion approach

Due to the scarcity of a benchmark dataset for rail defect generation, we explored alternative synthetic data generation methods as proposed in the literature.

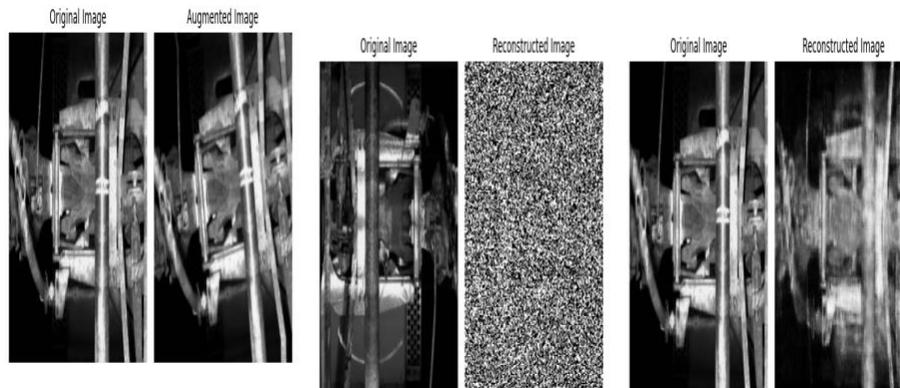

(a) Synthetic images generated by fixed-rule based geometric rotation (b) Output generated by GAN based approach (c) Synthetic images generated by proposed approach

Figure 11: Comparison of synthetic images generated by different approaches.

- Synthetic data generation by fixed-rule based geometric trans-formation We have demonstrated the results of employing geometric transformations on images in Fig. 11a. To achieve this, we employed ran-dom rotation techniques on the original images sourced from another CPR dataset, consisting of 1000 defective and 2000 normal images. In a previ-ous study [7], we employed this approach in the data pre-processing phase for differentiating between defective and non-defective images. Although this geometric augmentation led to improved accuracy in binary defect classification, we encountered challenges related to significant overfitting when employing CNN models.

Basically, the augmented images from fixed-rule-based geometric transfor-mations do not obtain substantial alterations in their patterns or textures. From Fig. 11a, it is visible that the augmented images lack authenticity in terms of their patterns and textures (e.g., there is no difference between the synthetic images other than the angle). This absence of realistic vari-ation contributed to a defect classifier in a preceding paper achieving high accuracy during training. However, during testing, the model experienced elevated loss and fluctuating performance. As a result, the application of geometric transformations did not sever the purpose of generalizing the rail defect classifier.

- Synthetic data generation by GAN model We extended our explo-ration to the CPR dataset by applying a GAN-based transformation. The result of this process is illustrated in Fig. 11b. We intended to utilize GAN to generate synthetic datasets from the given CPR dataset. However, we found that the GAN is limited to dealing with a relatively small sample size. In our case, the GAN model was unable to generate synthetic sam-ples. With only 50 samples, the GAN model could not capture the full range of diversity inherent in the target image dataset. In other words, the model collapsed and could not generate meaningful samples due to improper noise modeling.

  These findings underscored the dependency of the GAN model on avail-ability of a sufficiently large and diverse dataset for effective training. Therefore, in scenarios where defect data is infrequent and hard to collect GANs may not be suitable for generating synthetic defect samples due to their sensitivity to sample size.

- Synthetic data generation by the proposed VAE model The syn-thetic images generated by our proposed VAE approach are depicted in Fig. 11c. In this technique, we leveraged the latent space of the VAE to generate meaningful and diverse synthetic images from limited samples.

  Geometric transformations, as evidenced in Fig. 11b exhibit exact patterns and textures akin to original images; which usually leads to high accuracy but overfitted model. By contrast, the proposed approach by capturing the underlying distribution in the latent space of the image. The result in Fig. 11c illustrates that the proposed VAE-generated synthetic images exhibit realistic patterns and textures similar to those of the original im-ages. Such output subdues the limitations posed by the fixed-rule-based geometric transformation method where hardly any variation in texture and pattern is observed. Moreover, the weight decay-based regulariza-tion technique strengthened the proposed VAE approach to handle the overfitting issue.

  In summary, The comparative analysis of the three synthetic data gener-ation methods highlights the shortcomings of fixed-rule-based geometric transformations and the sensitivity of GAN-based transformation to sam-ple size limitations. In our subsequent evaluations, we demonstrate the superior performance of the defect classifier trained on the proposed VAE-generated synthetic data compared to the models trained on the other two approaches.

### 4.4 Evaluation of ViT defect Classifier

We generate a synthetic dataset from the CPR main dataset using Algorithm 1. This dataset comprises of 500 images, with 50 being original and the remaining being reconstructed images.

#### 4.4.1 Training

Based on the findings from our precedent research detailed in [8] and [6], we underscored the need for a reusable model for defect classification. Therefore, we trained a MobileViT model. MobileViT is a visual transformer, that integrates MobileNet CNN with the visual transformer approach to create an efficient and lightweight model. We performed 80:10:10 Training:Testing: Validation split. Consequently, we fine-tuned this model using pre-trained weights, employing 440 training images, 55 validation images, and 55 testing images, all distributed among 5 distinct classes derived from the CPR synthetic dataset.

#### 4.4.2 Classifier Performance

In Fig. 12, we illustrate the actual and predicted labels obtained from the classifier trained on a single batch of images representing one of the defect classes. This batch includes test samples from both synthetic and original images. No-tably, none of these test samples have been used in the training set the actual labels are predicted with high precision, yielding nearly 0% loss and 100% accu-racy. Such high accuracy holds significant importance in the context of digital twining diagnostic processes []. Therefore, these outcomes substantiate the ef-ficacy of leveraging generative AI to address issues related to limited defective samples and to build a accurate defect classifier.

Furthermore, we conducted an evaluation of the defect classifier trained on synthetic data for each class, focusing on precision, recall, and accuracy. The results showed that we achieved 98%-99% accuracy for precision, recall, and F1-score. The question of overfitting may arise when a model exhibits high accuracy with a relatively small sample size. Nevertheless, as evidenced by our qualitative output in Fig. 12, both the synthetic test defect samples and original test defect samples were accurately classified, providing further validation of the model's effectiveness

### 4.5 Discussion

Our comprehensive evaluation yields the following key findings:

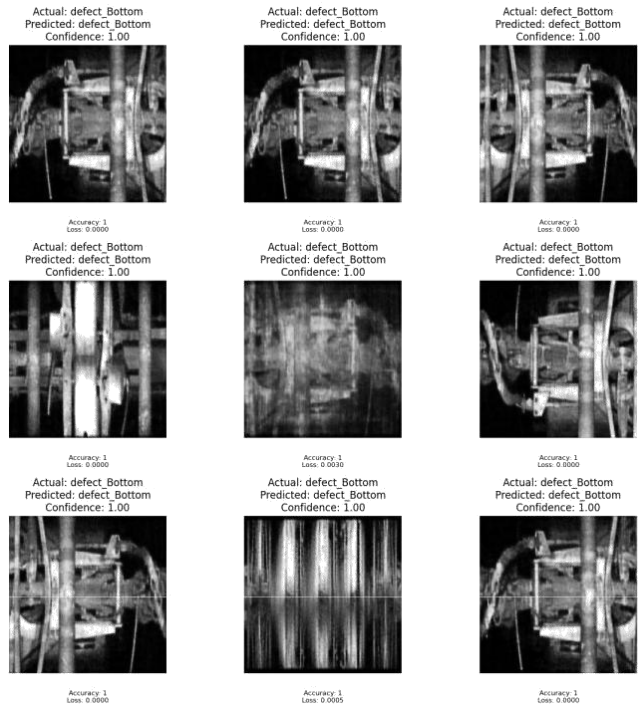

Figure 12: Actual and predicted labels are categorized by the trained classifier derived from a single batch of images, encompassing both synthetic and original images.

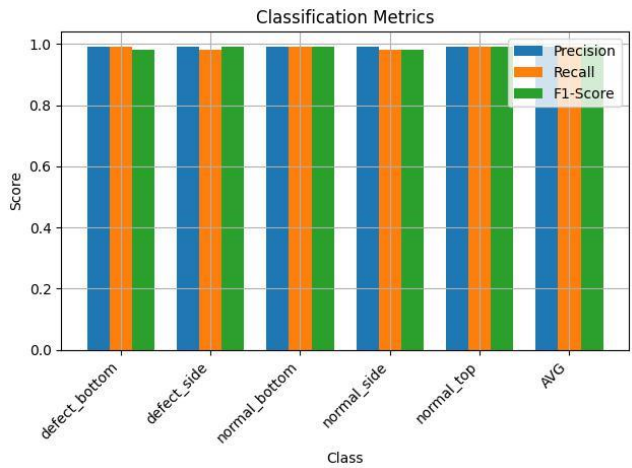

Figure 13: Comparison of classification metrics per class.

- Our synthetic data-trained defect classifier consistently exhibits high preci-sion, recall, and F1-score metrics, indicating its strong performance across all classes.

- The high precision values underscore the model's capacity to accurately identify and classify defects with minimal false positives. This precision is particularly vital in domains like railway quality control, where avoiding unnecessary interventions or alarms is paramount.

- The substantial recall scores signify the classifier's effectiveness in captur-ing a significant proportion of true positives. In safety-critical industries such as railways, this high recall rate implies the detection of a minimal number of defects that might otherwise ignored by the model.

- The achievement of a high F1-score highlights the model's efficiency, as it consistently maintains a balanced trade-off between precise defect identi-fication and false positive reduction.

In summary, the results from batch predictions and accuracy analysis show that our model performs effectively in classifying both synthetic and original defects, indicating good generalization.

## 5 Limitations and Future works

### 5.1 Limitations

Although our proposed approach outperformed conventional rule-based data augmentation strategies and demonstrated better performance compared to GAN-driven techniques in data quality and adaptability, there are a few limi-tations of this paper to mention.

First, while our approach performs better at generating precise replicas of defects, it may not be ideal for tasks demanding higher diversity in synthetic data, as our primary goal was ensuring accurate representation per class rather than promoting variation.

Second, the nuanced differences between certain defect types mean that oc-casionally, two almost identical defects might be reconstructed with varying quality.

Lastly, the approach was tailored for railway defect detection, and its appli-cability in other contexts remains to be seen.

### 5.2 Future Work

Although the model was primarily designed for railway defect detection, its po-tential for applications in other fields, where acquiring genuine training samples poses significant challenges. For instance,

- Aerospace and Aviation: Where simulated crashes or malfunctions are essential for safety research but physically creating such scenarios is prohibitively expensive and dangerous.

- Automotive Testing: Modern vehicles with advanced driver-assistance systems require vast amounts of data to ensure safety, and it's not feasible to induce real-world accidents or defects for data collection.

- Seismology: Creating datasets for earthquake prediction models would require waiting for real-world earthquakes, which are not only unpredictable but could be catastrophic.

- Medical Imaging for Rare Diseases: Obtaining sufficient samples for rare diseases without resorting to synthetic data is a major bottleneck.

In a nutshell, future studies could modify and fine-tune the proposed approach to produce more diverse synthetic data sets, particularly beneficial in scenarios where data generation is complex and challenging; and creativity and variation in data generation are crucial.

# 6 Conclusion

This study investigates the potential of integrating generative technique, such as VAE, within ICS to enhance learning and decision-making processes in areas where real-world data is scarce or difficult to procure. By employing a VAE-based approach for synthetic data generation, we demonstrated how ICS can emulate human-like analysis even with limited data. Our approach out-performed traditional rule-based data augmentation techniques and showcased improved robustness compared to GAN-based methods in terms of data quality and generalization. The application of our proposed approach in rail defect de-tection paves the way for further exploration and adoption of generative AI in ICS across various domains including health, climate, and heavy freights facing similar challenges.

# Declaration

## Authors Contribution

- Rahatara Ferdousi: Conceptualization, Methodology, Writing - Original Draft Preparation, Visualization, Validation, Software.

- Chunsheng Yang: Review & Data Collection, Validation, Project administration.

- M. Anwar Hossain: Writing-Review & Editing, Validation, Supervision.

- Fedwa Laamarti: Writing - Review & Editing.

- Abdulmotaleb El Saddik: Discussion, Review, Funding acquisition, Supervision.

## Funding

This research is supported in part by collaborative research funding from the National Program Office under the National Research Council of Canada's Arti-ficial Intelligence for Logistics Program. The project ID is AI4L-123. Also, the authors extend their appreciation to the researchers supporting project number: RSP2024R32, King Saud University, Riyadh, Saudi Arabia.

## Conflict of Interest

The authors declare that there are no conflicts of interest regarding the publication of this paper.

## Data Availability

We have signed a business agreement with the data provider. Accordingly, the datasets analyzed during the current study are not publicly available. However, we have consent for publishing the result.

# References

[1] Yandan Zheng and Luu Anh Tuan. A novel, cognitively inspired, unified graph-based multi-task framework for information extraction. Cognitive Computation, pages 1–10, 2023.

[2] Venkat N Gudivada, Sharath Pankanti, Guna Seetharaman, and Yu Zhang. Cognitive computing systems: Their potential and the future. Computer, 52(5):13–18, 2019.

[3] Tadahiro Taniguchi, Tomoaki Nakamura, Masahiro Suzuki, Ryo Kuniyasu, Kaede Hayashi, Akira Taniguchi, Takato Horii, and Takayuki Nagai. Neuro-serket: development of integrative cognitive system through the composi-tion of deep probabilistic generative models. New Generation Computing, 38:23–48, 2020.

[4] Y. Cao, Y. Wang, and W. Xu. Segmentation-based deep-learning approach for surface-defect detection. arXiv preprint arXiv:1903.08536, 2019.

[5] D. Tabernik, S. Sela, J. Skvarč, and D. Skočaj. Deep-learning-based computer vision system for surface-defect detection. In Computer Vision Systems: 12th International Conference, ICVS 2019, Thessaloniki, Greece, September 23–25, 2019, Proceedings 12, pages 490–500. Springer, 2019.

[6] Sara Ghaboura, Rahatara Ferdousi, Fedwa Laamarti, Chunsheng Yang, and Abdulmotaleb El Saddik. Digital twin for railway: A comprehensive survey. IEEE Access, 11:120237–120257, 2023.

[7] Rahatara Ferdousi, Fedwa Laamarti, Chunsheng Yang, and Abdulmotaleb El Saddik. Railtwin: a digital twin framework for railway. In 2022 IEEE 18th International Conference on Automation Science and Engineering (CASE), pages 1767–1772. IEEE, 2022.

[8] Chunsheng Yang, Rahatara Ferdousi, Abdulmotaleb El Saddik, Yifeng Li, Zheng Liu, and Min Liao. Lifetime learning-enabled modelling framework for digital twin. In 2022 IEEE 18th International Conference on Automa-tion Science and Engineering (CASE), pages 1761–1766. IEEE, 2022.

[9] S. Cui, H. Wang, M. Zhang, and X. Zhang. Defect classification on limited labeled samples with multiscale. Applied Intelligence, 51(6):3911–3925, 2021.

[10] Rajaa Alqudah, Amjed A Al-Mousa, Yazan Abu Hashyeh, and Omar Z Alzaibaq. A systemic comparison between using augmented data and syn-thetic data as means of enhancing wafermap defect classification. Comput-ers in Industry, 145:103809, 2023.

[11] Yun Xiao, Yameng Huang, Chenglong Li, Lei Liu, Aiwu Zhou, and Jin Tang. Lightweight multi-modal representation learning for rgb salient object detection. Cognitive Computation, pages 1–16, 2023.

[12] Synthetic data augmentation for surface defect detection and classification using deep learning.

[13] Defect-gan: High-fidelity defect synthesis for automated defect inspection.

[14] Improving vae based molecular representations for compound property pre-diction.

[15] X. He, Z. Chang, L. Zhang, H. Xu, H. Chen, and Z. Luo. A survey of defect detection applications based on generative adversarial networks. IEEE Access, 10:113493–113512, 2022.

[16] Mauro Castelli and Luca Manzoni. Generative models in artificial intelligence and their applications, 2022.

[17] Machine learning for synthetic data generation: A review. 2023.

[18] R. Lopez, P. Boyeau, N. Yosef, M. Jordan, and J. Regier. Decision-making with auto-encoding variational bayes. Advances in Neural Information Pro-cessing Systems, 33:5081–5092, 2020.

[19] Markus Endres, Asha Mannarapotta Venugopal, and Tung Son Tran. Syn-thetic data generation: a comparative study. In Proceedings of the 26th International Database Engineered Applications Symposium, pages 94–102, 2022.

[20] Lu Mi, Macheng Shen, and Jingzhao Zhang. A probe towards understanding gan and vae models. arXiv preprint arXiv:1812.05676, 2018.

[21] Lucas Pinheiro Cinelli, Matheus Araújo Marins, Eduardo Antúnio Barros da Silva, and Sérgio Lima Netto. Variational autoencoder. In Variational Methods for Machine Learning with Applications to Deep Networks, pages 111–149. Springer, 2021.

[22] Hugo Wai Leung Mak, Runze Han, and Hoover HF Yin. Application of variational autoencoder (vae) model and image processing approaches in game design. Sensors, 23(7):3457, 2023.

[23] Machine learning for synthetic data generation: A review.

[24] Teerath Kumar, Alessandra Mileo, Rob Brennan, and Malika Bendechache. Image data augmentation approaches: A comprehensive survey. 2023.

[25] Text data augmentation for deep learning. Journal of Big Data, 2021.

[26] Ruyu Wang, Sabrina Hoppe, Eduardo Monari, and Marco F. Huber. Defect transfer gan: Diverse defect synthesis for data augmentation. 2023.

[27] Gongjie Zhang, Kaiwen Cui, Tzu-Yi Hung, and Shijian Lu. Defect-gan: High-fidelity defect synthesis for automated defect inspection. 2021.

[28] Defect transfer gan: Diverse defect synthesis for data augmentation.

[29] Leveraging generative ai models for synthetic data generation in healthcare: Balancing research and privacy.

[30] Improving vae based molecular representations for compound property pre-diction. Journal of Cheminformatics, 2022.

[31] Ethem Alpaydin. Introduction to machine learning. MIT press, 2020.

[32] Hongbing Shang, Chuang Sun, Jinxin Liu, Xuefeng Chen, and Ruqiang Yan. Defect-aware transformer network for intelligent visual surface defect detection. Advanced Engineering Informatics, 55:101882, 2023.